\documentclass[letterpaper, 10 pt, conference]{ieeeconf}  

\IEEEoverridecommandlockouts                              

\usepackage{amsmath}
\usepackage{multirow}
\usepackage{graphicx}
\usepackage{bbold}

\title{\LARGE \bf Exploring the  trade off between human driving imitation and safety for traffic simulation}

\author{Yann Koeberle$^{1,2}$,Stefano Sabatini$^{2}$,Dzmitry Tsishkou$^{2}$ and Christophe Sabourin$^{1}$
\thanks{$^{1}$  {\tt\small Univ Paris Est Creteil, LISSI, F-77567 Lieusaint, France }}%
\thanks{$^{2}$ {\tt\small IoV team, Paris Research Center, Huawei Technologies France }}%
}
\begin{document}

\maketitle
\thispagestyle{empty}
\pagestyle{empty}

\begin{abstract}

Traffic simulation has gained a lot of interest for quantitative evaluation of self driving vehicles performance. 
In order for a simulator to be a valuable test bench, it is required that the driving policy animating each traffic agent in the scene acts as humans would do while maintaining  minimal safety guarantees. Learning the driving policies of traffic agents from recorded human driving data or through reinforcement learning seems to be an attractive solution for the generation of realistic and highly interactive traffic situations in uncontrolled intersections or roundabouts. In this work, we show that a trade-off exists between imitating human driving and maintaining safety when learning driving policies.
We do this by comparing how various Imitation learning and Reinforcement learning algorithms perform when applied to the driving task. We also propose a multi objective learning algorithm (MOPPO) that improves both objectives together. We test our driving policies on highly interactive driving scenarios extracted from INTERACTION Dataset to evaluate how human-like they behave.  
\end{abstract}

\section{INTRODUCTION}
Despite the huge amount of real driving data available, it is still very challenging to plan safe maneuvers for a Self Driving Vehicle SDV because it is impossible to expose it to all the possible situations it could face in reality \cite{filos2020autonomous}. Traffic simulation offers the possibility to augment driving experiences with synthetic episodes and has become a practical tool for self driving companies to design and evaluate algorithms for their vehicles \cite{scheel2022urban}. Simulation has a number of practical advantages over real test campaigns as the ability to control driving scenarios diversity, to explore near accident situations without endangering people, and most importantly, the ability to reproduce interactive driving episodes from the same initial settings \cite{rempe2021generating}. However, learning realistic driving policies for practical use cases in traffic simulation remains challenging because it assumes that traffic agents are able to interact with the SDV model under evaluation as human drivers would do. While traffic simulations on straight lanes of highways can safely be animated by heuristic based agents like IDM \cite{2000Treiber}, the interaction with other agents based on hard coded rules often does not look realistic in more complex locations like roundabouts or intersections. In contrast, learning based approaches \cite{filos2020autonomous,scheel2022urban, suo2021trafficsim,bhattacharyya_multi-agent_2018} offer more flexibility to adapt to many different situations leveraging large amounts of available human demonstrations.\\
\indent In order to assess the realism of traffic simulations, it is necessary to introduce metrics to measure  how similar traffic agents and  real human drivers are. Imitation metrics are defined with respect to real driving data and measure how close the simulated agent drives with respect to how a human driver would have driven in the same situation. Safety metrics are based on human prior knowledge of the task and track specific events such as collisions, off-road driving, traffic rules infractions or quantities such as jerk for comfort. Both type of metrics are necessary to describe performances of traffic agents and consequently, both need to be considered when building training objectives for the simulator driving policies. On one hand, optimizing imitation metrics is made possible with several approaches such as Imitation Learning (IL) \cite{codevilla2018endtoend}, or Adversarial Imitation Learning (AIL) \cite{ho2016generative}. However those approaches often perform poorly out of training distribution due to compounding errors  which can lead to crashes. Additionally, those algorithms often fail to exploit the true causal structure of the demonstration which limits their ability to adapt to new situations \cite{dehaan2019causal,codevilla2019exploring}.\\
\indent In contrast, it is also possible to design explicit reward functions based on prior knowledge of the task, with the objective to learn basic skills such as keeping safety distance margins and avoiding collisions. The driving policy can then be learned  with Reinforcement Learning (RL) by interacting with the environment following the guidance of this hand-crafted reward \cite{chen2019modelfree}. The main difficulty in this approach lies in potential miss specifications of the reward function \cite{knox2021reward} that could lead to unrealistic behaviours. Indeed, hand crafting a reward requires several potentially conflicting components  as jerk, safety margin, trajectory curvature, etc. and the designed reward only acts as a proxy of the true driving reward that is unknown. Consequently, maximizing the expected trajectory reward does not guarantee to recover human alike trajectories in every situations.\\
\indent In this this work, we quantitatively analyze the trade off that emerges between imitating  human driving behaviors and avoiding safety critical events such as collisions.

Our main contributions are:  
\begin{itemize}
\item analysing how different IL and RL algorithms balance safety and imitation performances of a driving policy on new real driving scenarios.
\item analysing how robust are the safety performances of a driving policy to changes of environment dynamics with the replacement of non interactive agents with interactive agents.
\item proposing a multi objective algorithm (MOPPO) that optimizes simultaneously safety and imitation performances. 
\end{itemize}

The remainder of this paper is the following: in the next Section \ref{sec:Related_Works}, we first give an overview of main approaches to learn either from demonstration either from prior task knowledge. Subsequently, in Section \ref{sec:Learning Driving Policies}, we formulate the problem of learning a driving policy for traffic simulation and we present our new MOPPO algorithm. We will then detail our experiments in Section \ref{sec:Experiments} and discuss the results in Section \ref{sec:Results}.

\section{RELATED WORKS}
\label{sec:Related_Works}
\subsection{Learning to drive from demonstrations}
The most typical approach to imitate an expert is called Behavior Cloning (BC) where the objective is to train a regression model to predict expert’s action given a training dataset \cite{Pomerleau1988ALVINNAA}. In this setting, during the policy execution, the collected observations will incrementally slightly shift from the observations used during training which violates the i.i.d assumption required for supervised learning. Ignoring this issue leads to poor performances both in theory and practice because as soon as the learner makes a mistake, it may encounter completely different observations than those seen during training leading to a compounding of errors \cite{ross2011reduction}. Concretely, driving policies learned with BC fail to plan safe trajectories in a wide spectrum of situations \cite{codevilla2019exploring}. To alleviate distributional shift, DAgger\cite{ross2011reduction} iteratively queries an interactive expert to avoid this kind of failures but such an interactive expert is not always available unless putting a human in the loop \cite{wu2021humanintheloop}. An alternative, is to minimize the divergence between the learner and the expert trajectory distribution interacting with a simulator using reinforcement learning techniques and a data driven reward. For example, AIL methods like GAIL \cite{ho2016generative} learns a driving policy that generates trajectories in a simulation environment. The collected reward represents how much the generated trajectory looks like an expert trajectory and it is obtained as a surrogate product of a discriminator trained to distinguish between expert and generated trajectories. It was shown in \cite{huang2022efficient} and \cite{behbahani2019learning} that AIL can learn driving policies that not only imitate the expert on its support, but also recover from new unseen situations yet collisions can still happen. AIL methods suffer from the fact that the discriminator ignores the causal structure of the interactions between the expert and the environment \cite{Zolna2020TaskRelevantAI}. Ignoring causality is particularly damaging because the feedback provided by the discriminator may exploit inappropriate features in the observation action pair, resulting in reasonable training performances but potentially low test performances \cite{filos2020autonomous}. Indeed a realistic and safe driving policy should in principle maintain a reasonable safety gap with respect to other agents which is not always the case as demonstrated by the remaining collisions of GAIL on straight highways \cite{kuefler2017imitating}.
To alleviate causal confusion, it is possible to augment the AIL reward with semantic components like a collision signal \cite{wang2021decision} either directly injected in the discriminator or added to the reward with a state based penalty \cite{bhattacharyya2019simulating}. 

\subsection{Learning to drive from human prior knowledge}

Commonly used autonomous driving datasets generally do not provide dangerous demonstrations and consequently approaches that are based on pure imitation struggle in handling safety critical situations. In contrast, reinforcement learning enables to learn a driving policy in pure simulated environment through a series of trials and errors. Facing in simulation a lot of safety critical situations for which the agent is penalized, considerably helps to learn how to recover from dangerous situations as shown by several works on complex urban driving scenarios \cite{chen2019modelfree,toghi2021learning}. However, human expert behaviours are often not provided for comparison and simulation environments are often fully virtual with hand designed road networks and traffic flows \cite{li2021metadrive}. It is consequently difficult to transfer RL policies to real world settings as no metric can guarantee that they would behave realistically. Additionally, RL policies performance can also suffer from a poorly designed reward function  which may cause the agent to mistakenly exploit the reward and stick to unexpected behaviours. As summarized in \cite{knox2021reward}, most of hand crafted reward functions consists of a weighted sum of bonus and penalties terms that quantifies progress to goal, safety margin, jerk, courtesy, etc. As a consequence, learning a driving policy that maximizes the hand crafted reward does not ensure the traffic agent to behave similarly to a human expert because the real reward that drives humans is complex and unknown. Practically, this can lead to differences, from moderate variations in speed distribution on road lanes which mainly depends on human preferences, up to undesired reactions such as risky commitments in intersections \cite{rana2021building}. Note that it is also possible to exploit expert demonstrations as suggested in \cite{huang2022efficient} that use uncertainty over cloned expert policies to regularize RL policy updates, or as in \cite{liu2021improved} where the expert demonstrations are artificially added to the replay buffer for off policy RL algorithms. Other works propose to include safety constraints in the driving policy by adding a rule based safety layer in case the learned policy action does not satisfy all safety requirements \cite{aksjonov2022safetycritical,vitelli2021safetynet}.

\section{Learning Driving Policies}
\label{sec:Learning Driving Policies}
\begin{figure*}[htb!]
    \begin{center}
        \includegraphics[width=0.9\textwidth]{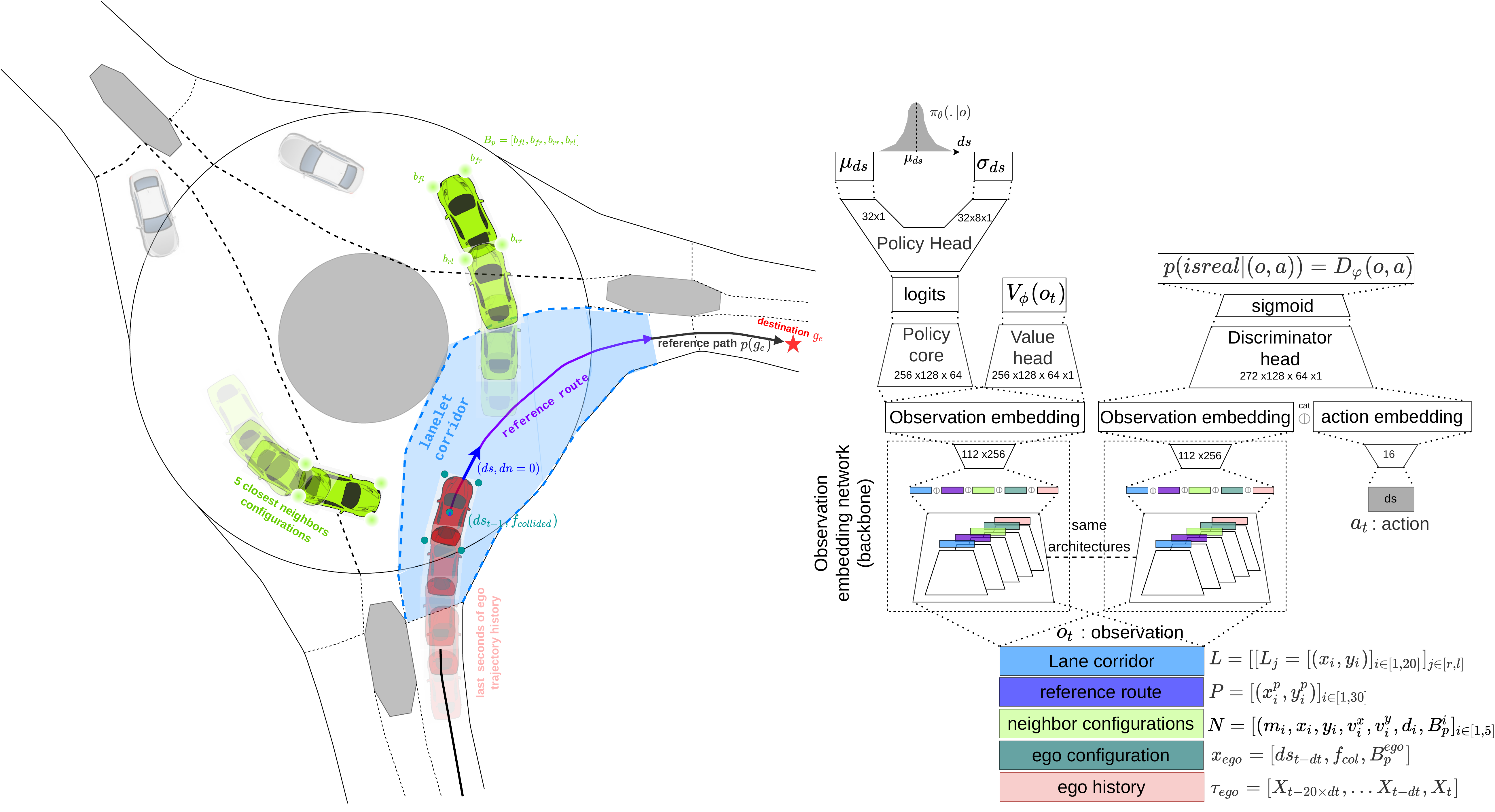} 
    \end{center}
\caption{On the left side, illustration of the components of the observation provided to the policy as well as the action generated by the policy and on the right side, architecture of the discriminator-actor-critic network used to train GAIL and PPO. }
\label{fig:Architecture_and_observation_action_space}
\end{figure*}

We start to formulate the problem of simulating traffic in Section \ref{subsec:Problem setting}. Section \ref{subsec:Learning policies} presents the main principles of RL and IL algorithms that are the starting point of our novel MOPPO algorithm presented in Section \ref{subsec:MOPPO}. We detail our observation and action spaces illustrated in Fig.\ref{fig:Architecture_and_observation_action_space} in Section \ref{subsec:Observation and action space} before introducing our neural network architectures in Section \ref{subsec:Driving policy architecture}.

\subsection{Problem setting}
\label{subsec:Problem setting}
The objective of a driving simulator is to generate driving episodes of a given temporal horizon $H$  based  on a driving scenario $ S=(\mathcal{R}_N,\mathcal{F},\rho_0,\mathcal{G}) $. The driving scenario is composed of a road network  $\mathcal{R}_N$, a traffic flow logic $\mathcal{F}$ that spawns traffic agents at specific locations according to an initial state distribution $\rho_0$ and that assigns to each agent a driving policy $\pi$ and a destination $g\in \mathcal{G}$. Note that destinations $\mathcal{G}$ can easily be extracted from the real road network $\mathcal{R}_N$  based on traffic priors. In this work we tackle the problem of learning driving policies for creating realistic traffic simulations. More specifically, we adopt a decentralized perspective over traffic simulation and learn a single actor driving policy in each driving episode\footnote{This approach is motivated by the fact that single agent learning is a first step toward multi agent learning that can turn highly unstable when initialized from scratch.}. For this reason, traffic agents are divided into two categories, the ones that learns called \textit{actors} and the ones that just animate the scene called \textit{workers} endowed with a fixed policy. Our object of interest is the driving episode  that is defined as the trajectory generated by the learning actor for $H$ steps. We leverage the formalism of Partially Observable Markov Decision Process (POMDP) $(\mathcal{S},\mathcal{O},\mathcal{A},\rho_0,\mathcal{T})$ to define a driving policy denoted $\pi: \mathcal{O},\mathcal{G} \to \mathcal{A} $ that matches an observation and a goal to an action. The environment evolves according to the transition dynamic $ \mathcal{T}:(s,a)\mapsto s'$ that depends on the actor driving policy action $a$ and the global traffic state $s$ encompassing actor and  workers individual states. The dynamic of traffic workers and their policies are directly included in the transition dynamic for facilitating notations. The observation provided to the driving  policy is generated based on an observer model $O:s \to o$ from the global traffic state $s$.

\subsection{Learning policies }
\label{subsec:Learning policies}
Since we formulate our problem as a POMDP, we introduce a reward function  $r:\mathcal{O}\times\mathcal{A}\times \mathcal{O}\to \mathbb{R}$ to define the objective of the driving task. Our goal will be to learn the parameters $\theta$ that maximize the expected return of the policy: $ max_{\theta} J(\pi_{\theta})$ where $ J(\pi_{\theta})=\mathbb{E}_{\tau \sim \pi_{\theta}}[G_0] $ and where $G_0=\sum_{t=0}^{H} \gamma^t.r(o_t,a_t,o_{t+1}) $ denotes the return of a trajectory collected by the actor driving policy $\pi_{\theta}$ in an episode of horizon H. In order to maximize the expected return we use Proximal Policy Optimization (PPO) \cite{schulman2017proximal} with the Generalized Advantage Estimator (GAE) \cite{schulman2018highdimensional}. The training objective $J^{PPO}(\theta)$ of the policy is expressed as follows $ max_{\theta}J^{PPO}(\theta)$:
\begin{multline*}
J^{PPO}(\theta) = \mathbb{E}_{(a_t,o_t)\sim \pi_{\theta_{old}}}[min(\mathcal{L}^{\pi}(\theta),\mathcal{L}^{\pi}_{clip}(\theta))] \\
\mathcal{L}^{\pi}(\theta) = \mu_t(\theta)\hat{A}_t^{GAE}(o_t,a_t) \\
\mathcal{L}^{\pi}_{clip}(\theta)=clip( \mu_t(\theta),1-\epsilon_{\pi},1+\epsilon_{\pi})\hat{A}_{t}^{GAE}(o_t,a_t) 
\end{multline*}
\label{Policy-objective}
where $\mu_t(\theta)=\frac{\pi_{\theta}(a_t|o_t)}{\pi_{old}(a_t|o_t)} $ denotes the likelihood ratio, $\epsilon_{\pi}$ the clipping threshold and $\pi_{old}$ the policy used for simulation rollouts. Note that we add an entropy bonus  $c_{e}\mathcal{H}[\pi_{\theta}](o_t) $ to the policy loss that we anneal during training with the entropy coefficient $c_{e}$ to favor exploration. In order to  estimate the generalized advantage estimator $\hat{A}_{t}^{GAE}(a_t,o_t)$\cite{schulman2017proximal} we also  regress the expected return with a value function $V_{\phi}$ according to the following clipped objective $min_{\phi}J^{value}(\phi)$:
\begin{multline*}
J^{value}(\phi)=\mathbb{E}_{o_t\sim \pi_{\theta_{old}}}[min(\mathcal{L}^{V}(\phi),\mathcal{L}^{V}_{clip}(\phi)] \\
\mathcal{L}^{V}(\phi) = (V_{\phi}(o_t)-R_t)^2 \\
\mathcal{L}^{V}_{clip}(\phi)=(clip(V_{\phi}(o_t),V_{\phi_{old}}(o_t)-\varepsilon,V_{\phi_{old}}(o_t)+\varepsilon)-R_t)^2
\end{multline*}
\label{Value-objective}\\
where $\varepsilon$ denotes a clipping threshold and $V_{\phi_{old}}$ the previous value function trained on trajectory samples of $\pi_{\theta_{old}}$. More details about the PPO set up is provided in \cite{schulman2017proximal}. To learn from prior knowledge of the driving task, we introduce a synthetic reward that is expressed as follows:
\begin{equation}
\label{synthetic reward}
r_s=r_{col}+\alpha r_{ds}
\end{equation}
It is composed of two weighted terms: one that penalizes collisions $r_{col}$ and one that encourages forward displacement along the path. The proposed reward model is the simplest way to specify the task without introducing too much prior bias \cite{knox2021reward} and avoid undesired side effects as sudden accelerations.\\
\indent While synthetic reward is sufficient to avoid most of the crashes, it is not rich enough to provide dense guidance for imitating human trajectories. Refining the signal  with hand crafted features for closer reconstruction of human trajectories on specific driving scenes would be highly inefficient and time consuming. In contrast to an hand-crafted reward, we introduce a data driven reward signal $r_D$ in order to guide the policy toward a human like behavior. We leverage adversarial imitation learning to learn the reward based on a discriminator $D_{\varphi}$ that is trained to distinguish expert and learner pairs $(o_t,a_t)$ as detailed in GAIL algorithm \cite{ho2016generative}. At the same time, the policy is trained to fool the discriminator maximizing the expected sum of the data driven reward. The objective of the discriminator $J^{Disc}(\varphi)$ constitutes a lower bound of  the Jensen Shannon divergence between policy and expert trajectory distributions and is expressed as follows $max_{\varphi}J^{Disc}(\varphi)$: 
\begin{multline} 
J^{Disc}(\varphi)=\\
\mathbb{E}_{(s,a)\sim \mathcal{B}_{\pi}} log(D_{\varphi}(s,a))+ \mathbb{E}_{(s,a)\sim \mathcal{B}_{e}} log(1-D_{\varphi}(s,a)) 
\end{multline} 
\label{Discriminator-objective}
Practically we alternate between optimizing the discriminator loss (\ref{Discriminator-objective}) and optimizing the PPO objective (\ref{Policy-objective}) computed with a surrogate reward coming from the discriminator. The discriminator output  $D_{\varphi}(o,a)$ represents the probability that $(o,a)$ was generated from an expert and enables to calculate the data-driven reward $r_D(s,a)=D_{\varphi}(s,a)$ as it is done in \cite{kostrikov2018discriminatoractorcritic}:
\begin{equation}
\label{data-driven reward}
\resizebox{.8\hsize}{!}{
 $r_D(o,a)=log(D_{\varphi}(o,a))-log(1-D_{\varphi}(o,a))$
 }
\end{equation}
Note that the signal $r_D(o,a)$ does not enable to recover the true expert reward \cite{Fu_AIRL} and directly depends on discriminator capacity to distinguish expert and learner. 

\subsection{Multi objective Proximal Policy Optimization}
\label{subsec:MOPPO}
\begin{figure}[htbp]
  \includegraphics[width=9.0cm]{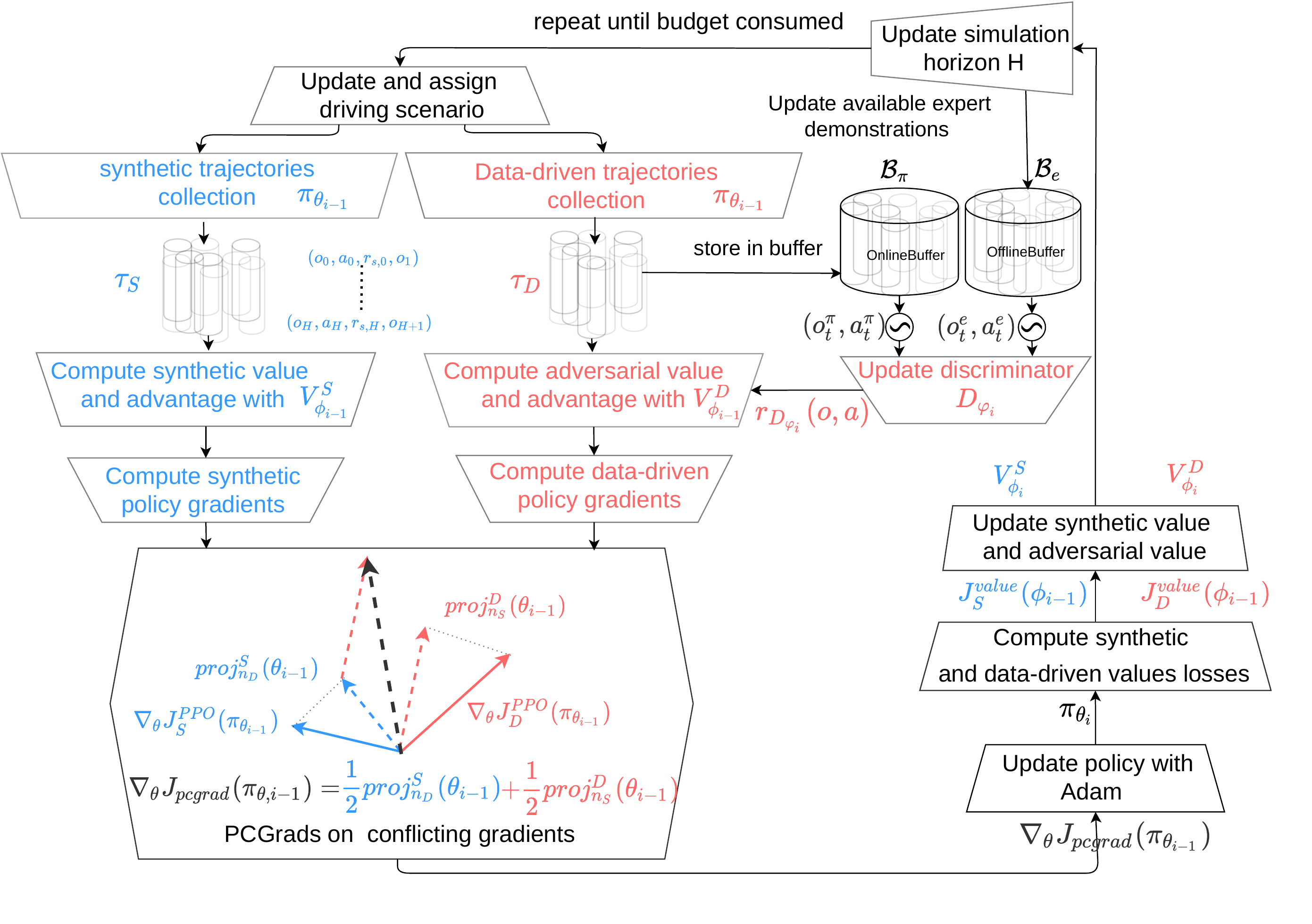}
    \caption{Multi Objective Proximal Policy Optimization (MOPPO): it computes two PPO objectives $J_S^{PPO},J_D^{PPO}$ with synthetic and data-driven rewards respectively and in case of conflicting gradients it applies PCGrad optimizer to compute the policy gradient. }
    \label{fig:MOPPO}
\end{figure}

In the previous Section \ref{subsec:Learning policies} we introduced two types of rewards that have different advantages for learning a driving policy. The synthetic reward $r_s$ enables to avoid catastrophic failures like collisions while encouraging the policy to reach its destination but the resulting trajectory can substantially differ from expert ones. By contrast, the data driven reward $r_D$ enables to extract specific features of human driving style but does not structure the behavior based on common sense priorities. It appears that combining both signals could be beneficial for learning safe and realistic policies and the most simple way would be to sum the two types of rewards. Despite their mutual contributions, the two rewards can also come into conflict either because the data-driven reward exploits inappropriate features of the observation or because the synthetic reward is inherently biased with respect to the true expert reward. Consequently it is necessary to find a policy parameters update that improves both synthetic and data-driven returns individually while solving the potential conflict that can arise between objectives.\\
\indent In order to tackle the above mentioned  challenge, we formulate a multi-objective optimization problem based on two separate PPO policy losses $J_S^{PPO}(\theta)$ and $J_D^{PPO}(\theta)$ computed on trajectories with the synthetic reward (\ref{synthetic reward}) and the data-driven reward (\ref{data-driven reward}) respectively. It has been shown that simply averaging the losses gradients does not always provides the correct solution to descend the multi-losses objective \cite{sener2019multitask}. Detrimental gradient interference appears when conflicting gradients\footnote{Gradients are called conflicting when they have negative cosine similarity \cite{yu2020gradient}.} coincide with high positive losses curvature and large differences in gradients magnitudes which is common for neural networks \cite{goodfellow2015qualitatively}. Consequently we propose to optimize the average loss based on the PCGrad optimizer \cite{yu2020gradient} which mitigates gradient interference by directly altering the gradients. If the two gradients $\nabla_{\theta}J_S^{PPO}(\theta)$ and $\nabla_{\theta}J_D^{PPO}(\theta)$ are conflicting as depicted in Fig.\ref{fig:MOPPO}, PCGrad projects each onto the normal plane of the other $proj^{S}_{n_D}(\theta),proj^{D}_{n_S}(\theta)$, preventing the interfering components of the gradient from being applied to the policy network.\\
\indent The full training procedures of our algorithm called $MOPPO$ is described on Fig.\ref{fig:MOPPO} and is decomposed in three main steps repeated until a training budget is consumed. In a first step, the most updated driving policy $\pi_{\theta}$ is used to collect massively in parallel, on multiple scenarios, two sets of trajectories $\tau_D$ and $\tau_S$ rewarded with the data driven and synthetic reward respectively. The data driven reward $r_D$ (\ref{data-driven reward}) is computed in a GAIL manner with a discriminator trained on recent policy trajectories $\tau_D$ and expert trajectories stored in a buffer $\mathcal{B}_{expert}$. In a second step, the GAE\cite{schulman2018highdimensional} is computed for both the data-driven and synthetic trajectories with separate value functions $V_{\phi_D}$  and $V_{\phi_S}$. This also enables to compute the two associate policy losses $J_{D}^{PPO}(\theta)$ and $J_{S}^{PPO}(\theta)$ with their respective gradients. Finally, the PCGrad optimizer sums the gradient projections in order to update the policy parameters $\theta$ in a suitable direction. Subsequently, the two value functions $V_{\phi_D}$  and $V_{\phi_S}$ are updated based on their respective returns before the procedure restarts. The whole training process of MOPPO also includes  horizon curriculum \cite{behbahani2019learning} that progressively increases the simulation horizon $H$ on both category of scenarios as the imitation performance of the policy improves. 

\subsection{Observation and action space }
\label{subsec:Observation and action space}
Each traffic agent interacts  with the simulator by observing the environment and taking actions. The actor driving policy generates the next action conditioned on the current observation $o_t$ and a goal $g$ assigned by the traffic flow. For each driving scenario $\mathcal{S}_i$, a reference goal $g_e^i$ enables to compute  a smooth path $p(g_e^i)$ connecting the actor to its desired destination thanks to a rule based routing module that operates on top of the policy. The path is computed based on the high definition map of the area and is enforced to match the path of the reference trajectory $\tau_e^i$ to avoid ambiguity during path search\footnote{On roundabouts for example, there are  several paths that conduct to an exit depending how many times the driver turn around the roundabout.}. In this work we focus on longitudinal control along the path $p(g_e^i)$ because it enables to handle intersections in a simple way since the agent is enforced to drive on the center lane that leads to its own goal.
As a consequence, we choose to output as action the longitudinal shift $ds$ along the path $p(g_e^i)$ provided to the policy. Regarding the observation, it is generated from the actor perspective and it contains several semantic components in a vectorized format as depicted in Fig.\ref{fig:Architecture_and_observation_action_space}. We provide a piece of the reference path $p(g_e^i)$ called reference route to indicate how the ego agent is expected to move longitudinally. The reference route $P=[(x_i^p,y_i^p)]_{i\in[1,30]}$ consists of a sequence of point coordinates relative to the actor sampled spatially uniformly every 50 cm in a 10 meter radius around the actor. In order to inform the agent about the local road topology, we provide left and right border points of the road corridor in which the agent is expected to move given the reference path. We leverage the lanelet2 map format \cite{poggenhans2018lanelet2} to extract the lane corridor $L=[ L_r=[(x_i,y_i)]_{i\in[1,20]}, L_l=[(x_i,y_i)]_{i\in[1,20]}]$ relative to actor position. To handle interactions, we provide the configurations $N=[(m_i,x_i,y_i,v^x_i,v^y_i,d_i,B_p^i]_{i\in [1,5]}$ of the 5 nearest neighbors around the actor. It contains a mask $m$ to indicate if the i-th neighbor exists, its position relative to the actor $x_i,y_i$, its  speed vector $v^x_i,v^y_i$, relative distances $d_i$ as well as its spatial extension with the four border points coordinates on front, rear, left and right side $B_p^i=[b_{fl}^{i},b_{fr}^{i},b_{rr}^{i},b_{rl}^{i}]$. Finally we also provide a component with ego agent information about itself as relative coordinates of its past 2 seconds trajectory $\tau_{ego}=[X_{t-20\times dt},...X_{t-dt},X_t]$, its last action $ds_{t-dt}$, a collision flag $f_{col}$, and its  spatial extension in terms of border points coordinates $B_p^{ego}=[b_{fl}^{ego},b_{fr}^{ego},b_{rr}^{ego},b_{rl}^{ego}]$. 

\subsection{Driving policy architecture }
\label{subsec:Driving policy architecture}
The actor driving policy $\pi_\theta$ is parametrized with a deep neural network represented on Fig. \ref{fig:Architecture_and_observation_action_space} with parameters denoted with $\theta$. As the observation is composed of multiple components, we first project each component separately with fully connected (FC) layers before concatenating the projections and applying another FC layer to obtain a global observation embedding vector. This embedding is then provided to a core policy network composed of two consecutive FC layers that generate logits vector. Logits will be used to compute the mean $\mu_{ds}$ and the standard deviation $\sigma_{ds}$ of a gaussian distribution that defines the next longitudinal shift $ds$. In addition to the policy, we also require a value function $V_{\phi}$ to estimate the advantage $ A_{GAE}^{\pi}(o_t,a_t) $ of action $a_t\sim\pi_{old}$ sampled from the policy distribution. To improve feature representation, the value head shares the observation embedding with the policy as depicted in Fig. \ref{fig:Architecture_and_observation_action_space} but the value loss is not back-propagated through the policy backbone to avoid training instabilities. For AIL algorithms like GAIL, we also introduce a discriminator $D_{\varphi}(o,a)$ to compute the surrogate reward from observation action pair $(o,a)$. The discriminator has the same backbone architecture as the policy and value function but does not share the same weights.

\section{Experiments}
We describe in Section \ref{subsec:Dataset and metrics} the dataset and the metrics we used for our analysis, then in Section \ref{subsec:Simulation}, we explain how we simulate other traffic agents and finally we introduce in Section \ref{subsec:Baselines} the baselines used in the experiments.
\label{sec:Experiments}
\begin{table*} 
  \caption{Imitation and safety metrics of driving policies evaluated on non interactive scenarios}
  \label{tab:safety and imitation}
  \centering
  \begin{tabular}{l|l|l l l| l l l l| l l }
    \hline
    \multirow{2}{*}{} &
      \multicolumn{1}{c}{RB} &
      \multicolumn{3}{|c|}{RL} &
      \multicolumn{4}{c|}{IL} &
      \multicolumn{2}{c}{MOPPO} \\
    \hline
    Metrics &  IDM & $PPO_M$ & $PPO_I$ & $PPO_{mix}$ & BC & $GAIL_M$ & $BCGAIL_M$ & $SGAIL_M$  & $MOPPO_{M}$ & $MOPPO_{mix}$\\
    \hline
    ADE-5(m) & 5.12 & 3.74 & 7.47 & 3.26 & 7.57 & 2.94 & 2.70 &  \textbf{2.17} & 2.45 & 2.25 \\  
    ADE-15(m) & 8.74 & 8.13 & 9.83 & 6.38 & 13.16  & 4.66  & 6.40 & 6.11 & \textbf{4.60} & 4.65 \\  
    CR(\%)  & 24 & 17 & 44 & 17 & 86 & 40 & 62 & 32 &  28 &  \textbf{10} \\  
    FCR(\%) & 11 & 10 & 35 & 10 & 64 & 30 & 37 & 10 & 20 & \textbf{8} \\ 
    \hline
  \end{tabular}
\end{table*}

\subsection{Dataset and metrics}
\label{subsec:Dataset and metrics}
We first introduce a common way to define each algorithm's training database so that fair comparisons between algorithms performances could be established. We define a training database as a set of real driving scenarios $\mathcal{D}=\{ \mathcal{S}_i \}_{[i\in[1,...,|\mathcal{D}|} $ that can be exploited either through simulation interactions by RL algorithms or directly by IL algorithms based on the associated expert demonstrations. The driving scenario used to train and evaluate  driving policies  are extracted from the Interaction Dataset driving scenes on the roundabout map called $DR\_DEU\_Roundabout\_OF$ \cite{zhan2019interaction}. The dataset contains numerous long horizon expert trajectories with high level of interaction between traffic agents which enables to learn collision avoidance strategy in challenging contexts. This is in contrast with previous works \cite{bhattacharyya_multi-agent_2018,boborzi2021learning}, that aim to learn driving policies on highways which limits the possibilities of critical situations with crashes. Another important aspect of the driving scenes present in the Interaction dataset, is the lanelet2 map format \cite{poggenhans2018lanelet2} which enables efficient observations extraction during parallel simulation roll-outs. In our case we extracted 200 scenarios for each of the following temporal horizons: 2.5, 5, 7.5, 10, 12.5 and 15 seconds which gives a total of 1200 training scenarios\footnote{This enables to extract each driving agent recorded in the scene even if it does not appear for a long time. }. The validation set is composed of 128 new real scenarios of 15 seconds in order to focus on long term simulation performances. In order to evaluate quantitatively the performance of the driving policy, we introduce two categories of metrics. Safety metrics deals with the collision experienced by the learning actor during the driving episode. We track both the rate of episodes with at least one collision (CR) and the rate of episodes where the collision occurs in a 60 degrees cone in front the the learning actor (FCR). Imitation metrics consists in the Average Distance Error in meters (ADE) with respect to recorded expert trajectories. We track both ADE-5 and ADE-15 that represent the error at 5 and 15 seconds respectively. While the rate of episodes with collision (CR) enables to understand if the policy can avoid catastrophic failures, the rate of episodes with front collision (FCR) is more appropriate to  understand how many times the policy is fully responsible for the crashes. Indeed some rear collisions are unavoidable due to non-reactive agents arriving behind the learning actor.

\subsection{Simulation}
\label{subsec:Simulation}
In order to learn driving policies on scenarios from the Interaction dataset, we designed our own simulator and leverage the RLlib framework to implement our algorithms \cite{liang2018rllib}. One crucial aspect of driving simulation is the dynamic of traffic workers: traffic agents that are not currently learning but animates the driving scene. We can run simulation in two different modes: either with interactive agents or with non reactive agents. In the non interactive setting, each traffic worker replays its original trajectory while the actor is learning\footnote{Since driving scenarios are extracted from real data, each traffic worker has a reference trajectory that can be replayed.}. In the interactive setting, traffic agents are controlled by an advanced version of the IDM model \cite{2000Treiber} that can handle intersections with minimal safety performances as attested in the next section. This situation occurs for example when IDM is about to enter the roundabout with a traffic agent on its left. Instead of exclusively considering front neighbors that lie in its own lane, our model considers all front neighbors in a 60 degrees cone in front of it that it transforms into virtual front neighbors by projecting their positions and their speeds on its own path. The acceleration of our advanced IDM agent is then regulated based on the position and speed of the closest front neighbor either real either virtual based on original equations of IDM model detailed in \cite{2000Treiber}. 

\subsection{Baselines}
\label{subsec:Baselines}
Since we aim to explore the trade off between human driving imitation and safe driving we compare the performances of several categories of algorithms that we list below:
\begin{itemize}
  \item \textbf{Rule Based}: we evaluate our advanced IDM model designed to also avoid collisions at intersections.
  \item \textbf{IL Baselines}: We trained a policy with Behavioural Cloning directly on demonstrations provided in the training set. We also introduce several variations of the GAIL algorithm. The basic GAIL was detailed in Section \ref{subsec:Learning policies} where the policy learns to imitate expert in simulation. We also trained a combination of BC and GAIL named BCGAIL that incorporates into the policy objective, imitation loss annealing as described in \cite{jena2020augmenting}. So far, $GAIL$ and $BCGAIL$ just exploit expert demonstrations through a data-driven reward (\ref{data-driven reward}) but a simple way to benefit from a synthetic feedback $r_S$ (\ref{synthetic reward}) would be to optimize the weighted sum of the rewards $r=0.5 r_S+0.5 r_D$ which leads to $SGAIL$ algorithm close to the algorithm proposed in \cite{wang2021decision}.
  \item \textbf{RL Baselines}: we also trained pure reinforcement learning baselines with the PPO algorithm as detailed in Section \ref{subsec:Learning policies}. We use a synthetic reward  $r_S=r_{col}+0.1r_{ds}$ that penalizes collisions by $r_d=-2$ when they occur otherwise $r_d=0$. The speed bonus $r_{ds}$ is computed as follows $r_{ds}=min(\frac{ds}{ds_{max}},1)$ based on the maximal longitudinal speed allowed $ ds_{max}=50km/hour$.
  \item \textbf{MOPPO}: we trained our new algorithm MOPPO detailed in Section \ref{subsec:MOPPO} that leverages both the synthetic reward (\ref{synthetic reward}) and the data-driven reward (\ref{data-driven reward}) from GAIL. 
\end{itemize}
In order to analyze the influence of the environment dynamic, we introduce three training variants. The first denoted with subscript $M$ indicates that the policy is trained with only replayed traffic workers, the second denoted with subscript $I$ indicates that the policy is trained with only interactive IDM workers and the last one with subscript ${mix}$ indicates that the policy is trained on each scenario twice, once with only replayed traffic workers and a second time with only IDM traffic workers.

\section{Results}
\label{sec:Results}
\subsection{Imitation and safety trade-off}

We first set up a series of experiments to analyze the trade-off between safety and human driving imitation. With this objective, we evaluate the driving policy learned with each of the algorithms presented in Section \ref{subsec:Baselines} on the scenarios belonging to our validation set. Since it is our interest to compute imitation metrics, the evaluation scenarios are populated with replay traffic workers as we do not have reference expert trajectories in traffic situations that evolve differently from the recordings.\\
\indent We start to compare performances of standard IL and RL algorithms to stress their respective benefits. We first observe that $GAIL$ exhibits better ADE-5 and ADE-15 than BC which quickly shifts due to compounding errors. The results of $BCGAIL$ show that  adding an imitation loss to GAIL helps to improve ADE-5 but tends to degrade long term performances. In contrast RL algorithms results show that $PPO$ mainly enables to reduce the number of collisions as shown by $PPO_M$ that reaches (CR:17\%). We notice that including interactive scenarios during training helps in improving imitation performances as demonstrated by the better ADE of $PPO_{mix}$ compared to $PPO_{M}$.
The trade off between imitation and safety arises clearly with $GAIL_M$ that reached lower ADE-5: (2.94m vs 3.74m) and ADE-15: (4.66m vs 8.13m) compared to $PPO_M$ whereas $PPO_M$ reached lower CR: (17\% vs 40\%) than $GAIL_M$.\\
\indent In order to combine benefits of both approaches, we evaluate our multi-objective algorithm MOPPO trained in two different settings. We observe that $MOPPO_M$ reached a decent balance : (ADE-5: 2.45m , ADE-15: 4.60m, CR: 28\%) but the collision rate remains higher than $PPO_M$. When we train MOPPO on a mixture of interactive and non interactive scenarios we observe that safety performances are considerably improved without degradation of imitation performances as shown by $MOPPO_{mix}$ results: (ADE-5: 2.25m , ADE-15: 4.65m, CR: 10\%). This variation enables to regularize the policy because it constantly introduces new traffic workers interactions that improve the collision avoidance strategy. We also note that MOPPO reaches a better trade-off compared to the simple combination of the rewards done by SGAIL. While $SGAIL$ imitates slightly better at short term, it exhibits bigger ADE-15 and CR than $MOPPO_M$ and $MOPPO_{mix}$ which attests that synthetic and data driven reward signals tend to conflict in the long term. This can be explained by the fact that the return $G_t^{D+S}$ obtained with the rewards sum $r=r_S+r_D$ introduces more variance in the policy gradient due to the covariance term of $\mathbb{V}[G_t^{D+S}]=\mathbb{V}[G_t^D]+\mathbb{V}[G_t^D]+Cov(G_t^D,G_t^S)$. Indeed, the data driven reward is getting increasingly more correlated with the synthetic reward due to their mutual influence on the trajectories collected by the policy along training.

\subsection{Safety robustness to interaction }
In this section, we analyse how robust are safety performances to changes of environment dynamics. Indeed, in previous section, we analysed safety metrics on new real scenarios with replay workers unable to react which is interesting as long as the actor stays close to expert original trajectory. If the actor tends to drift from expert original trajectory, it is interesting to analyse if it can yet adapt on the same scenarios except that replay workers are all replaced with interactive workers. As our advanced IDM model was specifically designed to interact with reasonable safety performances as attested by its low collision rate (CR:24\%), we expect that simulation with such traffic workers could generate valuable experiences for training the actor. Consequently we compare safety metrics (CR, FCR) summarized in table \ref{tab:safety robustness} of different algorithms on interactive (I) and non interactive setting (NI) on the validation set. We note that the best safety performances are obtained by $PPO_{mix}$ which reached the lowest rate of collisions (CR:21\%) on interactive scenarios and and by $MOPPO_{mix}$ which reached the lowest rate of collisions (CR:10\%) on non interactive scenarios. It appears that training with a mixture of interactive and non interactive driving scenarios is beneficial to the policy on both setting in evaluation as shown by $PPO_{mix}$ results compared to $PPO_M$ as well as $MOPPO_{mix}$ results compared to $MOPPO_M$. Note that despite $SGAIL$ improving safety performances of $GAIL$ on both settings, it has globally worse safety performances than $MOPPO_{M}$ and $MOPPO_{mix}$  which proves that combining rewards by addition is not as effective as a multi objective formulation.   

\begin{table}[htb!]
  \caption{Comparison of safety metrics of driving policies evaluated in interactive and non interactive settings}
  \label{tab:safety robustness}
  \centering
  \begin{tabular}{l|l|l}
    \hline
    \multicolumn{1}{c|}{} &
    \multicolumn{1}{c|}{CR(\%)} &
    \multicolumn{1}{c}{FCR(\%)} \\
    ALGORITHMS & NI/I & NI/I \\
    \hline
    BC &  86/82  &  64/59 \\
    IDM &  24/20  &  11/\textbf{9} \\
    $PPO_M$ &  17/40  &  10/24  \\
    $PPO_I$ &  44/28  &   35/12      \\
    $PPO_{mix}$ &  17/\textbf{21}  & 10/\textbf{9}   \\
    $GAIL_M$ &   40/78  &     30/27  \\
    $BCGAIL_M$ &  62/63  &  37/24   \\
    $SGAIL_M$ &  32/77  &   10/25  \\
    $MOPPO_{M}$ &  28/66  &  22/24   \\
    $MOPPO_{mix}$ &  \textbf{10}/32  &  \textbf{8}/12   \\  
    \hline
  \end{tabular}
\end{table}

\section{CONCLUSIONS}
In this works we analysed how different IL and RL algorithms  perform in terms of safety and imitation performances on real driving scenarios and compared their results to our new multi objective algorithm $MOPPO$. We noticed that $MOPPO_{mix}$ found the best balance between imitation and safety and even outperforms all algorithms in terms of imitation performances. We also analysed how robust are the safety performances of those algorithms to a change of environment dynamic by replacing replay traffic workers with IDM traffic workers. It appears that our algorithm successively generalizes its collision avoidance strategy with interactive agents. In future works, we should try to privilege safety performances to provide more guarantees about collision avoidance  while long tern imitation performances could be further improved by incorporating into $MOPPO_{mix}$ a fully differentiable policy gradient objective as introduced in \cite{scheel2022urban}.


\bibliography{RefYannISTC2022}
\bibliographystyle{IEEEtran}

\end{document}